\documentclass[conference]{IEEEtran}
\IEEEoverridecommandlockouts
\usepackage{cite}
\usepackage{amsmath,amssymb,amsfonts}
\usepackage{algorithmic}
\usepackage{graphicx}
\usepackage{textcomp}
\usepackage{xcolor}

\usepackage{subfloat}
\usepackage{float}

\usepackage{lipsum}
\usepackage{multirow}
\usepackage{graphicx}
\usepackage{textcomp}
\usepackage{xcolor}
\usepackage{makecell}
\usepackage{array}
\usepackage{threeparttable}
\usepackage{algorithm}
\usepackage{algorithmic}
\usepackage{amsthm}
\usepackage{booktabs}
\usepackage{bm}

\usepackage{tabu}
\usepackage{bbding}
\usepackage{rotating}
\usepackage{pifont}

\usepackage{colortbl}

\def\BibTeX{{\rm B\kern-.05em{\sc i\kern-.025em b}\kern-.08em
    T\kern-.1667em\lower.7ex\hbox{E}\kern-.125emX}}
\begin{document}

\title{FS-BAND: A Frequency-Sensitive Banding Detector\\
}

\author{Zijian Chen\textsuperscript{1}, Wei Sun\textsuperscript{1}, Zicheng Zhang\textsuperscript{1}, Ru Huang\textsuperscript{2}, Fangfang Lu\textsuperscript{3},\\ Xiongkuo Min\textsuperscript{1}, Guangtao Zhai\textsuperscript{1,*}, and Wenjun Zhang\textsuperscript{1}  \\
\textsuperscript{1}\textit{Institue of Image Communication and Information Processing, Shanghai Jiao Tong University, China}\\
\textsuperscript{2}\textit{School of Information Science \& Engineering, East China University of Science and Technology, China}\\
\textsuperscript{3}\textit{College of Computer Science and Technology, Shanghai University of Electric Power, China}\\
\thanks{*Corresponding author: Guangtao Zhai}
}

\maketitle

\begin{abstract}
Banding artifact, as known as staircase-like contour, is a common quality annoyance that happens in compression, transmission, etc. scenarios, which largely affects the user's quality of experience (QoE). The banding distortion typically appears as relatively small pixel-wise variations in smooth backgrounds, which is difficult to analyze in the spatial domain but easily reflected in the frequency domain.
In this paper, we thereby study the banding artifact from the frequency aspect and propose a no-reference banding detection model to capture and evaluate banding artifacts, called the \textbf{F}requency-\textbf{S}ensitive \textbf{BAN}ding \textbf{D}etector (FS-BAND).
 The proposed detector is able to generate a pixel-wise banding map with a perception correlated quality score. Experimental results show that the proposed FS-BAND method outperforms state-of-the-art image quality assessment (IQA) approaches with higher accuracy in banding classification task.
\end{abstract}


\section{Introduction}
\begin{figure}[!t]
	\centering
	{\includegraphics[width=0.48\textwidth]{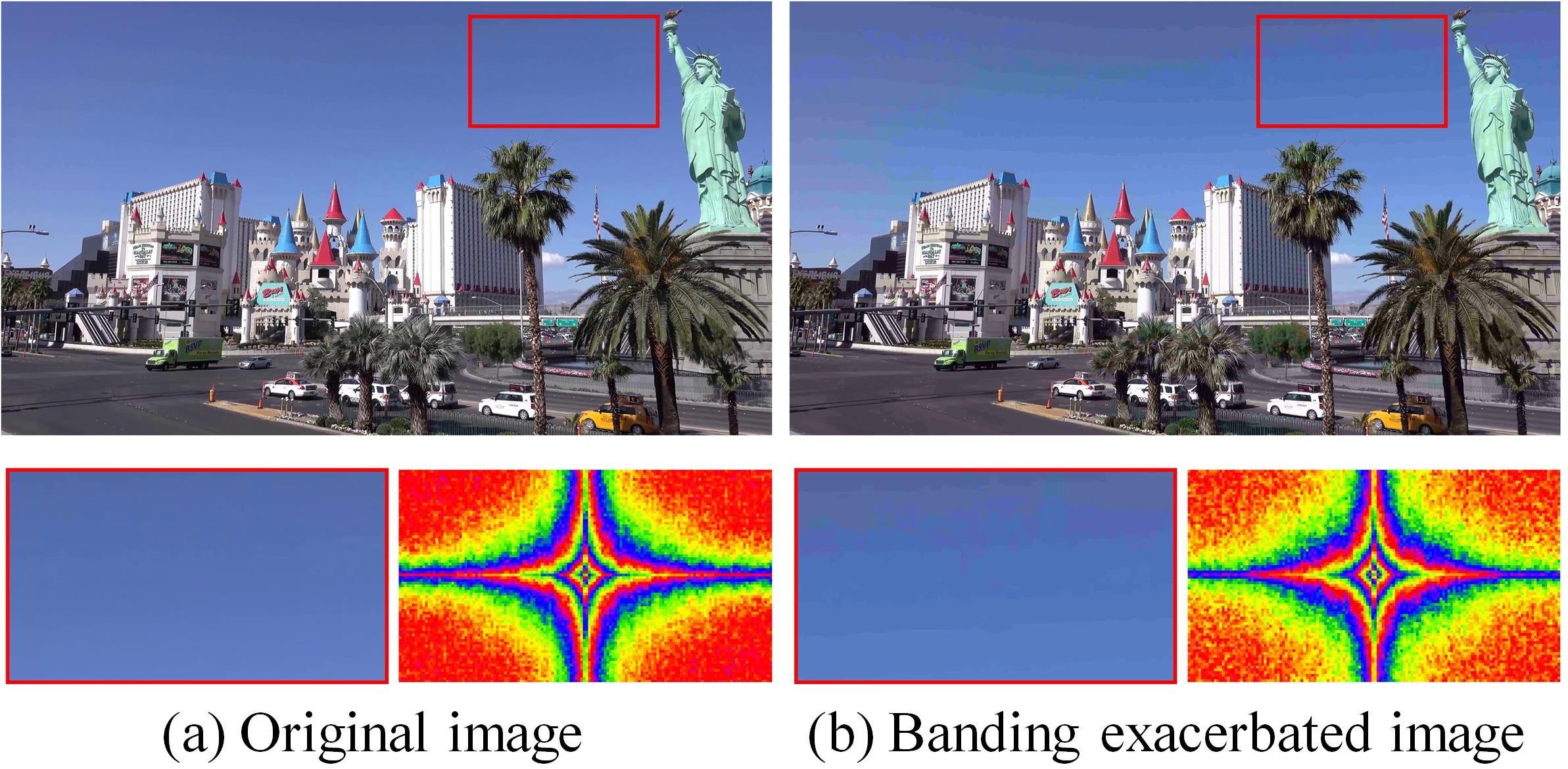}} 
	\caption{Illustration of the banding artifacts caused by bit-depth reduction. The figures in the bottom-right corner show the Fourier transformed spectrum of the boxed area, where the outer yellow scatters represent high-frequency information.}
	\label{intro}
\end{figure}

\fontdimen2\font=0.62ex
Banding artifacts are a kind of false contour distortion that is quite perceptible to the human eye. It usually takes on the appearance of annual rings, radiation circles, halos, or geographical contour lines and especially exist in the background regions (e.g. sky, water, and wall surface) where the color transition is not smooth enough. 
As shown in Fig. \ref{intro}, the banding artifacts in the sky are exacerbated by scaling down the bit-depth of the chroma channel. 
Normally, banding artifacts are triggered by the bit-depth reduction of luminance and chromaticity channels in the transcoded/codec process. Nearly all prominent video compression encoders, such as H.264/AVC \cite{wiegand2003overview}, VP9 \cite{mukherjee2013latest}, and H.265/HEVC \cite{sullivan2012overview} can introduce such artifacts more or less.
Besides, due to the limitation of shooting equipment and post-processing techniques, user generated content (UGC) are more likely to encounter such problem than professional generated content (PGC), which poses a huge challenge to both industry and academia.
Since the visual quality of image contents greatly affects the quality of experience (QoE) of end-users, it is highly desirable to design an effective banding detection and evaluation method, which aims to automatically detect the traces of false contours and predict the objective quality or severity of banding images that can be used to develop pre-processing or post-processing debanding algorithms and optimize the performance of streaming media application.

\begin{figure*}[!t]
\centering
{\includegraphics[width=1\textwidth]{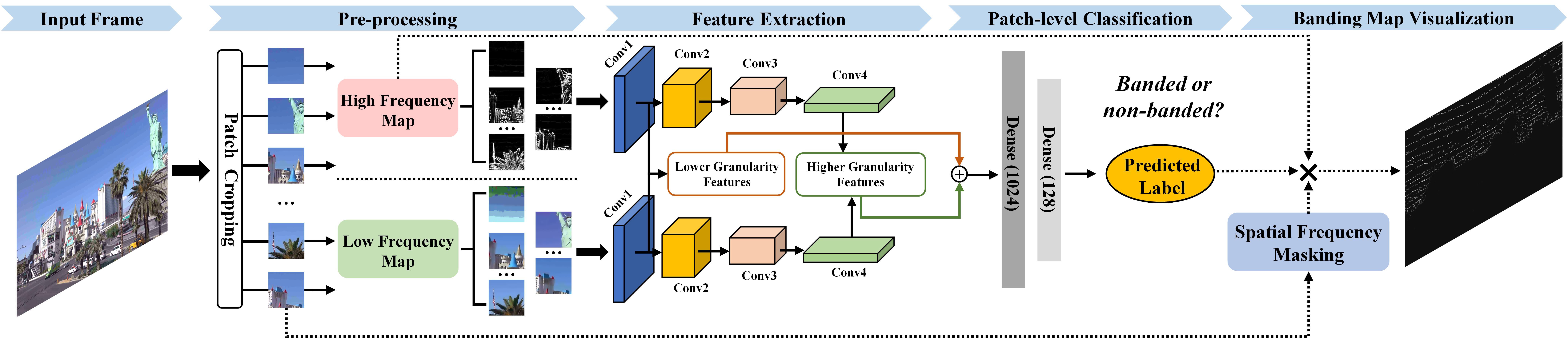}} 
\caption{Schematic overview of the proposed FS-BAND. Given a banding distorted image, it is first divided into patches. Then, the patch-level high-frequency map (HFM) and low-frequency map (LFM) are generated by Sobel operation and piece-wise smooth algorithm \cite{bar2006semi}, respectively. After that, a dual-CNN model is deployed to extract hierarchical features with different visual information and thus classify the patches into banded or non-banded. Lastly, a spatial frequency masking strategy is introduced to refine the banding map and calculate the image-level banding quality score.} \label{framework}
\end{figure*}

Early research on banding detection mainly focuses on false contour identification, which aims to find the wrong boundary rather than a ``true" region edge in the image. Authors in  \cite{lee2006two, huang2016understanding} utilized monotonicity or non-monotonicity features of local support regions including the gradient, contrast, variance, and entropy information to measure the loss of low-amplitude detail caused by banding. However, these works ignored the perceptual characteristics of the human visual system (HVS) and thus did not perform a good correlation with subjective tests. Another banding detection strategy is conducted at the pixel-level estimation and segmentation. Baugh \textit{et al.} \cite{baugh2014advanced} measured the severity of banding based on the number of a group of connected pixels with the same color. Wang \textit{et al.} \cite{wang2016perceptual} first detected uniform segments to find possible banding areas and further incorporated edge features (e.g. length and contrast) to capture false boundaries. Nevertheless, these kinds of methods are typically sensitive to edge noise and are computationally expensive, causing limited application in real-time scenarios.

Towards addressing these problems, Tu \textit{et al.} \cite{tu2020bband} presented a completely no-reference banding detection method, which combines various properties of HVS with a number of pre-processing steps to refine banding edge detection. Instead of regarding banding detection as a false edge detection problem, Tandon \textit{et al.} \cite{tandon2021cambi} heuristically utilized the effect of contrast sensitivity function (CSF) on banding visibility and its dependence on spatial frequency.
More recently, deep learning approaches have prevailed in various detection tasks. As the pioneering work, Kapoor \textit{et al.} \cite{kapoor2021capturing} developed an automated CNN-based banding detector for the first time, which is a simple two-stage algorithm and gives rise to devising other learning-based techniques. 

Our objective is to develop a fully blind banding predictor, which can detect and evaluate the banding artifacts independently and automatically. Fig. \ref{intro} shows a phenomenon that the spectrogram of the image exacerbated by banding artifacts contains more high-frequency information compared to the original image. In this regard, it could be an intuition that whether is it possible to use this frequency-sensitive characteristic to distinguish banding regions and thus design a banding detector.

In this paper, we propose a novel, no-reference banding model, dubbed the \textbf{F}requency-\textbf{S}ensitive \textbf{BAN}ding \textbf{D}etector (FS-BAND), based on frequency characteristics of banding artifacts and visual perception mechanisms, which utilizes a dual-branch CNN model to extract hierarchical banding-related feature representation from the high-frequency maps and low-frequency maps simultaneously. A spatial frequency masking strategy is introduced to refine the visibility of banding contours, and then combine with the detected banding map to generate subjectively consistent banding quality scores.

\section{The Proposed Banding Detector}
In this section, we describe the architecture of the proposed banding detector in detail, as shown in Fig. \ref{framework}.

\subsection{Frequency Map Generation}
As stated before, banding usually appear as high-frequency information in the smooth background, while humans perceive high-frequency texture regions and low-frequency plateau regions through different neural channels concurrently, and transfer the upper visual features into the cerebral cortex for final processing \cite{deyoe1988concurrent,zhang2022dual}. Inspired by this, we employ high-frequency maps (HFM) and low-frequency maps (LFM) as the deep learning network inputs, which represent the texture and structural information of the image respectively, to mimic the recognition mechanism of the human brain for better banding identification.

\noindent
\textbf{High-frequency Maps.} Since gradient has been widely used to represent edge information and has been confirmed beneficial to acquire high-frequency components with low computational cost \cite{tang2018full,gu2017evaluating,al2017multi}, we apply the isotropic Sobel operator to each patch for enhancing the details of banding artifacts. Given an input patch $\mathcal{I}$, the high-frequency map is calculated by
\begin{equation}
\mathcal{H}=\sqrt{\left( \mathcal{I} \ast \mathcal{S}_{x} \right)^{2}  +\left( \mathcal{I} \ast \mathcal{S}_{y} \right)^{2}  } ,
\end{equation}
where $\mathcal{S}_{x}$ and $\mathcal{S}_{y}$ are the horizontal and vertical isotropic Sobel operators, respectively. ``$\ast$" denotes the convolution operation.

\begin{figure}[!t]
	\centering
	{\includegraphics[width=0.48\textwidth]{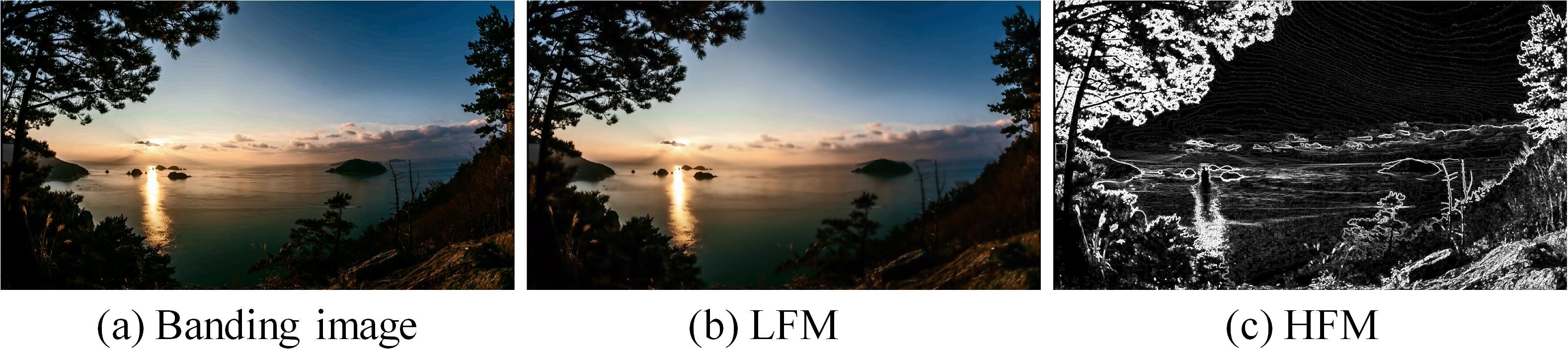}} 
	\caption{Exemplary frequency maps. The low-frequency map (LFM) is more smooth compared to the original image, while the high-frequency map (HFM) contains more sharp details. Zoom-in for better visualization.}
	\label{frequency-map}
\end{figure}

\noindent
\textbf{Low-Frequency Map:} 
To maintain the principal content of the image and filter out the influence of high-frequency information, we use the piece-wise smooth algorithm \cite{bar2006semi} to generate the low-frequency map by minimizing a function for image approximation recovery:

\begin{equation}
\mathcal{F} =\frac{1}{2} \int_{\Omega } (\mathcal{I} -\mathcal{L} )^{2}dP+\alpha \int_{\Omega \backslash E} |\nabla \mathcal{L} |^{2}dP+\beta \int_{E} d\sigma ,
\end{equation}
where $\mathcal{L}$ represents the low-frequency map, $\Omega$ and $E$ denotes the image domain and edge set, respectively. $P$ indicates the pixel and $\int_{E} d\sigma$ represents the total edge length. The coefficients $\alpha$ and $\beta$ are positive regularization constants. An example of frequency maps is shown in Fig. \ref{frequency-map}.

\subsection{Dual-CNN Model}
Considering that banding is a kind of high-frequency false edges surrounded by low-frequency contents, we reckon that detecting banding artifacts with the combination of both high- and low-frequency information can better distinguish whether there exist banding artifacts or other texture contents in the target region.

Therefore, we take both HFM and LFM as the inputs to classify the banding regions. As shown in Fig. \ref{framework}, the proposed model consists of two parallel branches. For each branch, we utilize Resnet50 \cite{he2016deep} as the backbone. Due to the fact that sharing parameters is extremely unfavorable for extracting low- and high-frequency features simultaneously, we thus deploy two networks that work independently and do not share parameters. Specifically, we incorporate the feature maps extracted from the first convolutional layer and the last layer of Resnet-50 as hierarchical visual features, which represent different visual information \cite{zeiler2014visualizing,ranjan2017hyperface} and can be used as predictive information to enhance the discrimination ability of the network for banding and non-banded regions.
Subsequently, the features extracted from two branches are concatenated first and reshaped into 128-dimensional vectors through two fully-connected layers, which is further followed with the sigmoid activation function to output the final predicted label, namely banded or non-banded.
Note that sharing parameters is extremely unfavorable for extracting low- and high-frequency features simultaneously, we thereby deploy two branches that work independently and do not share parameters. 
Here, we leverage the binary cross entropy as the loss function:
\begin{equation}
Loss =-[y\cdot \log(p(x))+(1-y) \cdot \log \left( 1-p(x)\right)],
\end{equation}
where $p(x)$ denotes the output probability and $y$ is the ground truth that appears as 0 or 1.

\subsection{Banding Detection and Evaluation}
To better quantify the severity of the banding, it is necessary to generate a quality score for the entire banding image. Since the visibility of banding edge is also affected by content, we further consider the effect of spatially varying content information on the local quality of human perception. As a consequence, we introduce the spatial frequency masking strategy to determine the weighting factor for the detected banding regions in each patch adaptively and thus obtain the image-level banding severity score while refining the visibility of banding artifacts.

\subsubsection{Spatial Frequency Masking}
The spatial frequency is defined as the activity level of an image, which establishes a filter-bank based on the visual stimulus and is in accordance with HVS \cite{li2008multifocus}. In this paper, we propose to apply spatial frequency as an effective contrast criterion to banding measurement. Specifically, given an image of size $I_W \times I_H$, divided into $N \times N$ patches, where $I_W$ and $I_H$ denote the number of columns and rows respectively. The column ($CF_k$) and row ($RF_k$) frequencies of the image patches are given by
\begin{align}
	CF_{k}&=\sqrt{\frac{1}{N^2} \sum^{N}_{x=2} \sum^{N}_{y=1} \left( I\left( x,y\right)  -I\left( x-1,y\right)  \right)^{2}  } \\
	RF_{k}&=\sqrt{\frac{1}{N^2} \sum^{N}_{x=1} \sum^{N}_{y=2} \left( I\left( x,y\right)  -I\left( x,y-1\right)  \right)^{2}  } ,
\end{align}
where $I(x,y)$ is the pixel value of the image patch. Then, the resulting spatial frequency is $SF_{k}=\sqrt{CF^{2}_{k}+RF^{2}_{k}} $, where $k$ is the serial number of patches. Since most banding regions are likely to have large contrast including edges and textures, which should be assigned greater weights than the smooth and blurred areas. Following Kazemi \textit{et al.} \cite{kazemi2022multifocus}, we set a threshold value to distinguish these regions, which is defined as the average spatial frequency of image patches:
\begin{equation}
	\epsilon =\frac{N^{2}}{I_{W}I_{H}} \sum^{I_{W}I_{H}/N^{2}}_{k=1} SF_{k},
\end{equation}
Accordingly, we design the following weight function, which measures the visibility of $k$-$th$ banding artifacts:
\begin{equation}
w_{k}=\begin{cases}1&|SF_{k}|\leq \epsilon \\ 1+\left( |SF_{k}|-\epsilon \right)^{\gamma }/N &|SF_{k}|>\epsilon, \end{cases} 
\end{equation}
where $\gamma$ is the scaling factor that is used to tune the enhancement of banding edges. In this work, the value of $\gamma$ is set to $1.5$ empirically.

\begin{figure*}[!t]
\centering
{\includegraphics[width=1\textwidth]{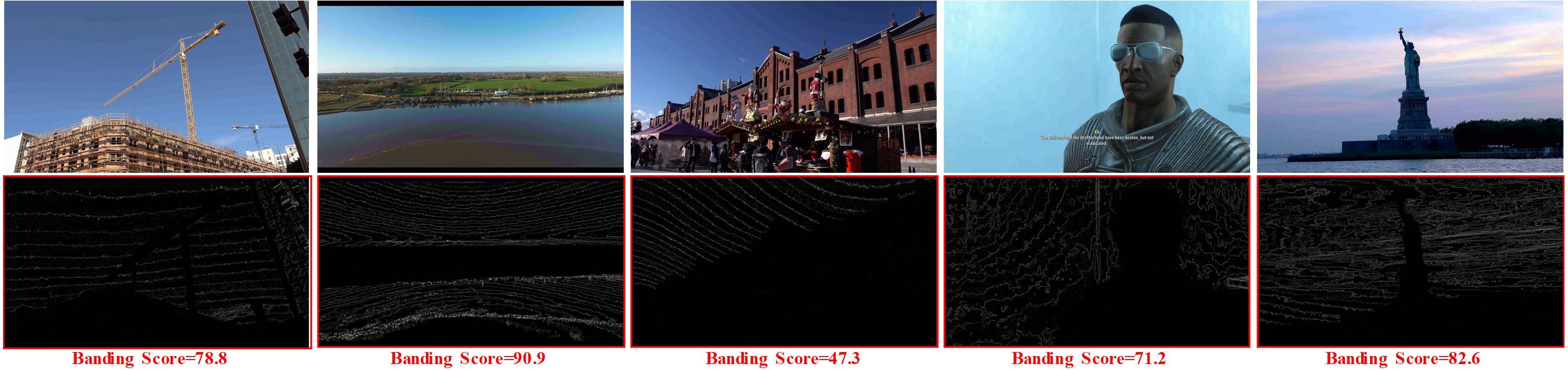}} 
\caption{Exemplary banding maps and their corresponding predicted quality scores.} \label{example}
\end{figure*}

\subsubsection{Building a Banding Metric}
The visibility of banding artifacts depends on the combination of multiple visual mechanisms.
In this paper, we propose a simple but effective product model for attribute integration at each predicted banding patch to obtain the entire banding map (BM):
\begin{equation}
	\mathrm{BM}_{k}(i,j) =w_{k}\cdot \widehat{P_{k}} \cdot |\mathrm{HFM}_{k} \left( i,j\right)  |,
\end{equation}
where $\widehat{P_{k}}$ denotes the predicted label of $k$-th patch and $w_k$ is the weight parameter that scales the visibility of measured contours, \textit{i.e.}, gradient magnitude of the high-frequency map, $|\mathrm{HFM}_{k} \left( i,j\right)|$ at region $(i,j)$.
Furthermore, inspired by previous psycho-visual findings that the QoE of observers is dominated by those regions having poor quality \cite{ghadiyaram2017no,tu2020bband}, we thereby leverage the worst $p\%$ percentile visual pooling to calculate an average banding score from the generated BM, where $p$ is set to $80$ in this experiment. As a result, the perceptual score of the overall banding image is defined as
\begin{equation}
	\mathcal{Q} \left( \mathcal{I} \right)  =\frac{1}{M} \cdot \frac{1}{|\mathcal{T}_{p\% } |} \sum^{M}_{k=1} \sum\nolimits_{\left( i,j\right)  \in \mathcal{T}_{p\% } } \mathrm{BM}_{k} \left( i,j\right) , 
\end{equation}
where $M$ is the total number of patches in image $\mathcal{I}$. $\mathcal{T}_{p\% }$ denotes the index set of the top $p\%$ non-zero pixel-wise value contained in $k$-th patch of the BM.

\section{Experiments}
In this work, we compare the proposed FS-BAND model with existing IQA approaches on the task of patch banding classification using the database released in \cite{kapoor2021capturing} to evaluate the performance of the proposed method. 
To the best of our knowledge, this is the only publicly available banding artifacts database. Concretely, we select 5 full-reference (FR) IQA methods: PSNR, SSIM \cite{wang2004image}, MS-SSIM \cite{wang2003multiscale}, LPIPS \cite{zhang2018unreasonable}, VMAF\textsubscript{BA} \cite{krasula2022banding} and 9 no-reference (NR) IQA methods including BRISQUE \cite{mittal2012no}, NIQE \cite{mittal2012making}, NIMA \cite{talebi2018nima}, DBCNN \cite{zhang2018blind}, HyperIQA \cite{su2020blindly}, StairIQA \cite{sun2023blind}, BBAND \cite{tu2020bband}, CAMBI \cite{tandon2021cambi}, and DBI \cite{kapoor2021capturing} as baselines. During the training stage, we randomly split the training and testing set into 8:2. Adam optimizer with a learning ratio set as 1e-3 is used. The batch size is set as 32. 
To make a fair comparison, all models are performed on a Windows PC with Intel i5-9300HF CPU 2.4GHz, 16GB RAM, and NVIDIA GeForce GTX 1660Ti 6G GPU. For patch-level banding classification, we utilize the area under the receiver operating characteristics (AUROC), the area under the precision-recall curve (AUPRC), and accuracy as the classification performance metrics. Since all IQA methods produce scalar values only while failing in classifying banding regions directly, we adopt a thresholding strategy to convert the single quality value into binary classification results as \cite{kapoor2021capturing} does, and a half-interval search algorithm \cite{bentley1975multidimensional} is employed to find the optimal threshold value that can generate the best classification result. 

\begin{table}
\centering
\caption{Experimental Results of Patch-Level Banding Classification. Accuracy and Speed are Reported in the Form of Maximum Testing Accuracy and Execution Time in Seconds per Image Patch. The Best Result is Highlighted}
\label{exp}
\renewcommand\arraystretch{0.9}
\begin{tabular}{c|l|cccc}
\toprule[0.75pt]
Type&Model&AUROC$\uparrow$ &AUPRC$\uparrow $ &Accuracy$\uparrow $ &Speed$\downarrow$ \\
\midrule[0.5pt]
\multirow{5}{*}{FR}&PSNR&0.0585&0.0543&25.34\%&\textbf{0.0071}\\
&SSIM \cite{wang2004image}&0.2421&0.2417&48.63\%&0.0091\\
&MS-SSIM \cite{wang2003multiscale}&0.2543&0.2674&52.63\%&0.0153\\
&LPIPS \cite{zhang2018unreasonable}&0.6571&0.6428&71.33\%&0.0098\\
&VMAF\textsubscript{BA} \cite{krasula2022banding}&0.2955&0.2746&47.11\%&0.0173\\
\midrule[0.5pt]
\multirow{10}{*}{NR}&BRISQUE \cite{mittal2012no}&0.2638&0.3163&56.31\%&0.0295\\
&NIQE \cite{mittal2012making}&0.1627&0.2134&44.15\%&0.0412\\
&NIMA \cite{talebi2018nima}&0.2853&0.2767&46.32\%&0.0125\\
&DBCNN \cite{zhang2018blind}&0.7435&0.7327&74.92\%&0.0147\\
&HyperIQA \cite{su2020blindly}&0.7652&0.7626&78.66\%&0.1758\\
&StairIQA \cite{sun2023blind}&0.7117&0.6933&67.34\%&0.1053\\
&BBAND \cite{tu2020bband}&0.3322&0.3175&45.72\%&0.1002\\
&CAMBI \cite{tandon2021cambi}&0.1553&0.1468&28.63\%&0.0087\\
&DBI \cite{kapoor2021capturing}&0.9442&0.9461&91.23\%&0.0231\\
&FS-BAND\textsubscript{(ours)}&\textbf{0.9872}&\textbf{0.9833}&\textbf{96.41\%}&0.0252\\
\bottomrule[0.75pt]
\end{tabular}
\end{table}

Table \ref{exp} reports the performance comparison results on the patch-level banding classification task. We can observe that our FS-BAND yields the best overall performance in terms of AUROC, AUPRC, and accuracy. It is shown that most general FR IQA and NR IQA models perform poorly on the patch-level banding classification task, indicating that the current approaches are not sensitive to banding distortion. However, the performance of banding IQA method BBAND, CAMBI, and VMAF\textsubscript{BA} is surprisingly poor compared with other methods, which shows their vulnerability in identifying local banding artifacts from texture regions and are not suitable for patch-level banding identification. In addition, we investigate the computational complexity in terms of execution time per image patch. It can be observed that except for those traditional FR IQA models, our method achieves comparable speed in patch-level banding classification, which determines the prediction efficiency of the subsequent image-level quality assessment, making it a favorable choice in time-constrained scenarios. Fig. \ref{example} gives a visualization of the generated banding maps with predicted quality scores, which shows the subjective consistent between predicted scores and banding severity. We also explore the effectiveness of our model's design philosophy in Table \ref{exp2}. It demonstrates the effectiveness of the dual-branch scheme and validate our hypothesis that the high-frequency texture information contained in HFM and the low-frequency background information contained in LFM are crucial to enhance the capacity of discernment for
banding artifacts.

\begin{table}
\centering
\caption{Ablation Study of FS-BAND. `SB' and `DB' Indicate that We Deploy Single Branch and Dual Branch Input, Respectively, while `HFM', `LFM', and `I' Denotes that We Use the High-Frequency Map, Low-Frequency Map,\\ and Banding Image as Inputs.}
\label{exp2}
\begin{tabular}{l|cccc}
\toprule[0.75pt]
Model&AUROC$\uparrow$&AUPRC$\uparrow$&Accuracy$\uparrow$&Speed$\downarrow$ \\
\midrule[0.5pt]
SB-HFM&0.8489&0.8322&84.26\%&\textbf{0.0219}\\
SB-LFM&0.8123&0.8137&82.21\%&0.0228\\
SB-I&0.9129&0.9058&89.21\%&0.0235\\
DB-HFM&0.8548&0.8617&84.35\%&0.0246\\
DB-LFM&0.8251&0.8233&82.57\%&0.0271\\
FS-BAND&\textbf{0.9527}&\textbf{0.9534}&\textbf{94.18\%}&0.0252\\
\bottomrule[0.75pt]
\end{tabular}
\end{table}

\section{Conclusion}
In this paper, we proposed a novel banding detector using the frequency characteristic of banding artifacts, which models the banding as high-frequency artifacts that contained in the low-frequency
smoothing region. A dual-branch CNN is devised to extract hierarchical features to classify the banding regions, upon which we introduce the spatial frequency masking to refine and compute an overall banding score. Experimental results show that our proposed method outperforms the baseline algorithms significantly in patch-level banding classification. Ablation studies verified that the proposed dual-branch structure and frequency-sensitive characteristics are effective.
Further extensions can be made toward a wider scope of machine-oriented banding detection, which includes more exploration of deep learning techniques and perceptual objects.

\bibliographystyle{IEEEbib.bst}


\end{document}